\documentclass[11pt,letterpaper]{article}
\usepackage[letterpaper]{geometry}
\usepackage{acl2012}
\usepackage{amsmath}  
\usepackage{times}
\usepackage{txfonts}  
\usepackage{latexsym}
\usepackage{multirow}
\usepackage{color}
\usepackage{url}
\usepackage{tikz}
\usepackage{tikz-qtree}
\usepackage{comment}
\makeatletter
\newcommand{\@BIBLABEL}{\@emptybiblabel}
\newcommand{\@emptybiblabel}[1]{}
\makeatother
\usepackage[hidelinks]{hyperref}
\title{Unsupervised Grammar Induction with Depth-bounded PCFG}

\author{
Lifeng Jin \\
Department of Linguistics \\
The Ohio State University \\
{\tt jin.544@osu.edu} 
\And
Finale Doshi-Velez \\
Harvard University \\
{\tt finale@seas.harvard.edu} 
\And
Timothy Miller \\
Boston Children's Hospital \& \\
Harvard Medical School \\
{\tt timothy.miller@childrens.harvard.edu} 
\AND
William Schuler \\
Department of Linguistics \\
The Ohio State University \\
{\tt schuler@ling.osu.edu} 
\And
Lane Schwartz \\
Department of Linguistics \\
University of Illinois at Urbana-Champaign \\
{\tt lanes@illinois.edu} }

\date{}

\def\diag{\mathrm{diag}}
\def\multinom{\mathrm{Multinom}}
\def\dirich{\mathrm{Dirichlet}}
\newcommand{\indicator}[1]{\llbracket #1 \rrbracket}
\newcommand{\prob}[1]{{\sf P}(#1)}
\newcommand{\condprob}[2]{{\sf P}(#1 \mid #2)}
\newcommand{\modelcondprob}[3]{{\sf P}_{#1}(#2 \mid #3)}
\def\defeq{\smash{\overset{\mathrm{def}}{=}}}
\def\dec{-}
\def\inc{+}

\def\CV{\mathbf{C}}

\begin{document}
\maketitle
\begin{abstract}
There has been recent interest in applying cognitively or empirically motivated bounds on recursion depth to limit the search space of grammar induction models \cite{Ponvert2011,Noji2016d,Shain2016}.
This work extends this depth-bounding approach to probabilistic context-free grammar induction (DB-PCFG), which has a smaller parameter space than hierarchical sequence models, and therefore more fully exploits the space reductions of depth-bounding.
Results for this model on grammar acquisition from transcribed child-directed speech and newswire text exceed or are competitive with those of other models when evaluated on parse accuracy.
Moreover, grammars acquired from this model demonstrate a consistent use of category labels, something which has not been demonstrated by other acquisition models.
\end{abstract}

\section{Introduction}

Grammar acquisition or grammar induction \cite{Carroll1992} has been of interest to linguists and cognitive scientists for decades. This task is interesting because a well-performing acquisition model can serve as a good baseline 
for examining factors of grounding \cite{zettlemoyercollins05,kwiatkowskietal10}, or as a piece of evidence \cite{Clark2001,Zuidema2003} about the Distributional Hypothesis \cite{harris54} against the poverty of the stimulus \cite{Chomsky1965}.
Unfortunately, previous attempts at inducing unbounded context-free grammars \cite{Johnson2007,Liang2009} converged to weak modes of a very multimodal distribution of grammars.
There has been recent interest in applying cognitively or empirically motivated bounds on recursion depth to limit the search space of grammar induction models \cite{Ponvert2011,Noji2016d,Shain2016}.
\newcite{Ponvert2011} and \newcite{Shain2016} in particular report benefits for depth bounds on grammar acquisition using hierarchical sequence models, but either without the capacity to learn full grammar rules (e.g.~that a noun phrase may consist of a noun phrase followed by a prepositional phrase), or 
with a very large parameter space that may offset the gains of depth-bounding.
This work extends the depth-bounding approach to directly induce probabilistic context-free grammars,%
\footnote{\url{https://github.com/lifengjin/db-pcfg}}
which have a smaller parameter space than hierarchical sequence models, and therefore arguably make better use of the space reductions of depth-bounding.
This approach employs a procedure for deriving a sequence model from a PCFG  \cite{VanSchijndel2013}, developed in the context of a supervised learning model, and adapts it to an unsupervised setting.

Results for this model on grammar acquisition from transcribed child-directed speech and newswire text exceed or are competitive with those of other models when evaluated on parse accuracy.
Moreover, grammars acquired from this model demonstrate a consistent use of category labels, as shown in a noun phrase discovery task, something which has not been demonstrated by other acquisition models.

\section{Related work}

This paper describes a Bayesian Dirichlet model of depth-bounded probabilistic context-free grammar (PCFG) induction.
Bayesian Dirichlet models have been applied to the related area of latent variable PCFG induction \cite{Johnson2007,Liang2009}, in which subtypes of categories like noun phrases and verb phrases are induced on a given tree structure.
The model described in this paper is given only words and not only induces categories for constituents but also tree structures.

There are a wide variety of approaches to grammar induction outside the Bayesian modeling paradigm.
The CCL system \cite{Seginer2007a} uses deterministic scoring systems to generate bracketed output of raw text.
UPPARSE \cite{Ponvert2011} uses a cascade of HMM chunkers to produce syntactic structures.
BMMM+DMV \cite{Christodoulopoulos} combines an unsupervised part-of-speech (POS) tagger BMMM and an unsupervised dependency grammar inducer DMV \cite{Klein2004a}.
The BMMM+DMV system alternates between phases of inducing POS tags and inducing dependency structures.
A large amount work \cite{Klein2002,Klein2004a,Bod2006,Berg-Kirkpatrick2010,Gillenwater2011PosteriorParsing,HeaddenIii2009,Bisk,Scicluna,Jiang,Han2017DependencyData} has been on grammar induction with input annotated with POS tags, mostly for dependency grammar induction. Although POS tags can also be induced, this separate induction has been criticized \cite{Pate2016} for missing an opportunity to leverage information learned in grammar induction to estimate POS tags.
Moreover, most of these models explore a search space that includes syntactic analyses that may be extensively center embedded and therefore are unlikely to be produced by human speakers.
Unlike most of these approaches, the model described in this paper uses cognitively motivated bounds on the depth of human recursive processing to constrain its search of possible trees for input sentences.

Some previous work uses depth bounds in the form of sequence models \cite{Ponvert2011,Shain2016}, but these either do not produce complete phrase structure grammars \cite{Ponvert2011} or do so at the expense of large parameter sets \cite{Shain2016}.
Other work implements depth bounds on left-corner configurations of dependency grammars \cite{Noji2016d}, but the use of a dependency grammar makes the system impractical for addressing questions of how category types such as noun phrases may be learned.
Unlike these, the model described in this paper induces a PCFG directly and then bounds it with a model-to-model transform, which yields a smaller space of learnable parameters and directly models the acquisition of category types as labels.

Some induction models learn semantic grammars from text annotated with semantic predicates \cite{zettlemoyercollins05,%
kwiatkowskietal12}.
There is evidence humans use semantic bootstrapping during grammar acquisition \cite{naigles90}, but these models typically rely on a set of pre-defined universals, such as combinators \cite{steedman00}, which simplify the induction task.
In order to help address the question of whether such universals are indeed necessary for grammar induction, the model described in this paper does not assume any strong universals except independently motivated limits on working memory.

\section{Background}
\label{sect:bkgd}

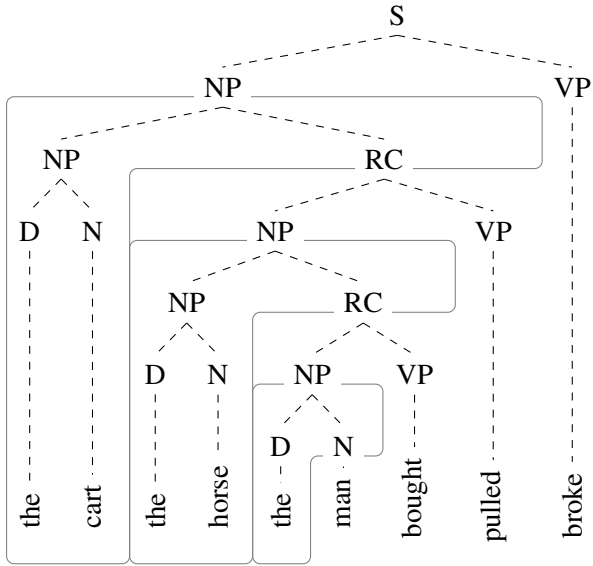
\begin{figure}
\begin{tikzpicture}[baseline,x=2.5em,y=2\baselineskip,level distance=2\baselineskip,sibling distance=3mm]
\tikzstyle{every node}=[anchor=mid,fill=white]
\tikzset{edge from parent path={(\tikzparentnode.south) edge[dashed] (\tikzchildnode.north)},frontier/.style={distance from root=6.5cm}}
\draw[gray,   rounded corners=1mm] (-5.4,-1) --     (2.0,-1) -- 
                                                   (2.0,-2) --
                                            (-3.7,-2) --
                                            (-3.7,-7.5) --
                                   (-5.4,-7.5) -- cycle;
\draw[gray,   rounded corners=1mm] (-3.7,-3) --    (.8,-3) -- 
                                                   (.8,-4) --
                                            (-2.0,-4) --
                                            (-2.0,-7.5) --
                                   (-3.7,-7.5) -- cycle;
\draw[gray,   rounded corners=1mm] (-2.0,-5) --    (-.2,-5) -- 
                                                  (-.2,-6) --
                                            (-1.2,-6) --
                                            (-1.2,-7.5) --
                                   (-2.0,-7.5) -- cycle;
\Tree[.S [.NP [.NP [.D \rotatebox{90}{\!\!\!the} ]
                   [.N \rotatebox{90}{\!\!\!cart} ] ]
              [.RC [.NP [.NP [.D \rotatebox{90}{\!\!\!the} ]
                             [.N \rotatebox{90}{\!\!\!horse} ] ]
                        [.RC [.NP [.D \rotatebox{90}{\!\!\!the} ]
                                  [.N \rotatebox{90}{\!\!\!man} ] ]
                             [.VP \rotatebox{90}{\!\!\!\!\!bought} ] ] ]
                   [.VP \rotatebox{90}{\!\!\!\!\!\!pulled} ] ] ]
         [.VP \rotatebox{90}{\!\!\!\!\!broke} ] ]
\end{tikzpicture}
\caption{Derivation fragments before the word {\em man} in a left-corner traversal of the sentence {\em The cart the horse the man bought pulled broke.}}
\label{fig:conncomps}
\end{figure}

Like \newcite{Noji2016d} and \newcite{Shain2016}, the model described in this paper defines bounding depth in terms of memory elements required in a left-corner parse.
\def\av{a}
\def\bv{b}
A left-corner parser \cite{rosenkrantzlewis70,johnsonlaird83,abneyjohnson91,resnik92lc} uses a stack of memory elements to store derivation fragments during incremental processing.
Each derivation fragment represents a disjoint connected component of phrase structure~$\av/\bv$ consisting of a top sign~$\av$ lacking a bottom sign~$\bv$ yet to come.
For example, Figure~\ref{fig:conncomps} shows the derivation fragments in a traversal of a phrase structure tree for the sentence {\em The cart the horse the man bought pulled broke.}
Immediately before processing the word {\em man}, the traversal has recognized three fragments of tree structure: two from category NP to category RC (covering {\em the cart} and {\em the horse}) and one from category NP to category N (covering {\em the}).
\def\dv{d}
\def\DV{D}
Derivation fragments at every time step are numbered top-down by depth~$\dv$ to a maximum depth of $\DV$.
A left-corner parser requires more derivation fragments --- and thus more memory ---  to process center-embedded constructions than to process left- or right-embedded constructions, consistent with observations that center embedding is more difficult for humans to process \cite{chomskymiller63,millerisard64}.
Grammar acquisition models \cite{Noji2016d,Shain2016} then restrict this memory to some low bound: e.g.\ two derivation fragments.

\def\wv{w}
\def\tv{t}
For sequences of observed word tokens~$\wv_{\tv}$
\def\TV{T}
for time steps~$\tv \in \{1..\TV\}$,
\def\qv{q}
sequence models like \newcite{Ponvert2011} and \newcite{Shain2016} hypothesize sequences of hidden states~$\qv_\tv$.
\def\fv{f}
\def\jv{j}
\def\pv{p}
Models like \newcite{Shain2016} implement bounded grammar rules as depth bounds on a hierarchical sequence model implementation of a left-corner parser, using random variables within each hidden state~$\qv_\tv$ for:
\begin{enumerate}
\item preterminal labels~$\pv_\tv$ and labels of top and bottom signs,~$\av^\dv_\tv$ and~$\bv^\dv_\tv$, of derivation fragments at each depth level~$\dv$ (which correspond to left and right children in tree structure), and 
\item boolean variables for decisions to `fork out'~$\fv_\tv$ and `join in'~$\jv_\tv$ derivation fragments (in \newcite{johnsonlaird83} terms, to {\em shift} with or without {\em match} and to {\em predict} with or without {\em match}).
\end{enumerate}
\def\MV{\mathbf{M}}
\def\dvm{\bar{d}}
Probabilities from these distributions are then multiplied together to define a transition model~$\MV$ over hidden states:
\begin{subequations}
\begin{align}
\MV_{[\qv_{\tv-1},\qv_\tv]}
  =      \,&\condprob{ \qv_\tv }{ \qv_{\tv-1} } \\
  \defeq   &\condprob{ \fv_\tv \, \pv_\tv \, \jv_\tv \, \av^{1..\DV}_\tv \, \bv^{1..\DV}_\tv }{ \qv_{\tv-1} } \\
  =      \,&\condprob{ \fv_\tv          }{ \qv_{\tv-1} } \notag\\
  \cdot  \,&\condprob{ \pv_\tv          }{ \qv_{\tv-1} \, \fv_\tv } \notag\\
  \cdot  \,&\condprob{ \jv_\tv          }{ \qv_{\tv-1} \, \fv_\tv \, \pv_\tv } \notag\\
  \cdot  \,&\condprob{ \av^{1..\DV}_\tv }{ \qv_{\tv-1} \, \fv_\tv \, \pv_\tv \, \jv_\tv } \notag\\
  \cdot  \,&\condprob{ \bv^{1..\DV}_\tv }{ \qv_{\tv-1} \, \fv_\tv \, \pv_\tv \, \jv_\tv \, \av^{1..\DV}_\tv }
\end{align}
\end{subequations}

For example, just after the word {\em horse} is recognized in Figure~\ref{fig:conncomps}, the parser store contains two derivation fragments yielding {\em the cart} and {\em the horse}, both with top category NP and bottom category RC.
The parser then decides to fork out the next word {\em the} based on the bottom category RC of the last derivation fragment on the store.
Then the parser generates a preterminal category D for this word based on this fork decision and the bottom category of the last derivation fragment on the store.
Then the parser decides {\em not} to join the resulting D directly to the RC above it, based on these fork and preterminal decisions and the bottom category of the store.
Finally the parser generates NP and N as the top and bottom categories of a new derivation fragment yielding just the new word {\em the} based on all these previous decisions, resulting in the store state shown in the figure.

The model over the fork decision ({\em shift} with or without {\em match}) is defined in terms of a depth-specific sub-model~$\theta_{\mathrm{F},\dvm}$, where $\bot$ is an empty derivation fragment and $\dvm$ is the depth of the deepest non-empty derivation fragment at time step~$\tv-1$:
\begin{multline}
\!\!\!\!\condprob{ \fv_\tv }{ \qv_{\tv\dec1} }
 \defeq\, \modelcondprob{\theta_{\mathrm{F},\dvm}} {\fv_\tv} {\bv^{\dvm}_{\tv\dec1}}; \ \ \dvm \!=\! \max_{\dv}\{\bv^{\dv}_{\tv\dec1} {\neq} \bot\} \!
\label{eqn:f}
\end{multline}

The model over the preterminal category label is then conditioned on this fork decision.
When there is no fork, the preterminal category label is deterministically linked to the category label of the bottom sign of the deepest derivation fragment at the previous time step (using $\indicator{\phi}$ as a deterministic indicator function, equal to one when $\phi$ is true and zero otherwise).
When there is a fork, the preterminal category label is defined in terms of a depth-specific sub-model~$\theta_{\mathrm{P},\dvm}$:%
\footnote{
Here, again, $\dvm \!=\! \max_{\dv}\{\bv^{\dv}_{\tv\dec1} \!\neq\! \bot\}$.
}
\begin{equation}
\condprob{ \pv_\tv }{ \qv_{\tv\dec1} \, \fv_\tv } \defeq
\begin{cases}
 \indicator{ \pv_\tv{=}\bv^{\dvm}_{\tv\dec1}}
   & \text{if } \fv_\tv=0
\\
 \modelcondprob{\theta_{\mathrm{P},\dvm}} {\pv_\tv} {\bv^{\dvm}_{\tv\dec1}}
   & \text{if } \fv_\tv=1
\end{cases}
\end{equation}

The model over the join decision ({\em predict} with or without {\em match}) is also defined in terms of a depth-specific sub-model~$\theta_{\mathrm{J},\dvm}$ with parameters depending on the outcome of the fork decision:%
\footnote{
Again, $\dvm \!=\! \max_{\dv}\{\bv^{\dv}_{\tv\dec1} \!\neq\! \bot\}$.
}
\begin{align}
\!\!\!\condprob{ \jv_\tv }{ \qv_{\tv\dec1}\ \fv_\tv\ \pv_\tv }
 \defeq\, &\begin{cases}
   \modelcondprob{ \theta_{\mathrm{J},\dvm}      }{ \jv_\tv }{ {\bv^{\dvm\dec1}_{\tv\dec1}} \ \av^{\dvm}_{\tv\dec1} } & \!\!\text{if } \fv_\tv {=} 0
   \\
   \modelcondprob{ \theta_{\mathrm{J},\dvm\inc1} }{ \jv_\tv }{ {\bv^{\dvm}_{\tv\dec1}}        \ \pv_\tv            } & \!\!\text{if } \fv_\tv {=} 1
   \end{cases}
\end{align}
%

Decisions about the top categories of derivation fragments $\av^{1..D}_t$ (which correspond to left siblings in tree structures) are decomposed into fork- and join-specific cases.
When there is a join, the top category of the deepest derivation fragment deterministically depends on the corresponding value at the previous time step.
When there is no join, the top category is defined in terms of a depth-specific sub-model:%
\footnote{
Here $\phi_{\dvm} = \indicator{\av^{1..\dvm}_\tv\!=\!\av^{1..\dvm}_{\tv\dec1}}$,
     $\psi_{\dvm} = \indicator{\av^{\dvm\inc1..\DV}_\tv\!=\!\bot}$,
and again, $\dvm \!=\! \max_{\dv
}\{\bv^{\dv}_{\tv\dec1} \!\neq\! \bot\}$.
}
\begin{align}
\modelcondprob{\theta_\mathrm{A}} {\av^{1..\DV}_\tv} {\qv_{\tv\dec1}\ \fv_\tv\ \pv_\tv\ \jv_\tv} \defeq 
\notag\\
  &\hspace{-4cm}\begin{cases}
    \phi_{\dvm\dec2} \cdot
    \indicator{a^{\dvm\dec1}_t\!=\!a^{\dvm\dec1}_{t\dec1}}   %
    \!\!\!\!&\cdot\ \psi_{\dvm\inc0}
      \ \ \text{ if } f_t,j_t\!=\!0,1
    \\
    \phi_{\dvm\dec1} \cdot
    \modelcondprob{ \theta_{\mathrm{A},\dvm} }{ a^{\dvm}_t }{ b^{\dvm\dec1}_{t\dec1}\ a^{\dvm}_{t\dec1} } 
    \!\!\!\!&\cdot\ \psi_{\dvm\inc1}
      \ \ \text{ if } f_t,j_t\!=\!0,0
    \\
    \phi_{\dvm\dec1} \cdot
    \indicator{a^{\dvm}_t\!=\!a^{\dvm}_{t\dec1}}   %
    \!\!\!\!&\cdot\ \psi_{\dvm\inc1}
      \ \ \text{ if } f_t,j_t\!=\!1,1
    \\
    \phi_{\dvm\dec0} \cdot
    \modelcondprob{ \theta_{\mathrm{A},\dvm\inc1} }{ a^{\dvm\inc1}_t }{ b^{\dvm}_{t\dec1}\ p_t }  
    \!\!\!\!&\cdot\ \psi_{\dvm\inc2}
      \ \ \text{ if } f_t,j_t\!=\!1,0
  \end{cases}
\end{align}
%

Decisions about the bottom categories $b^{1..D}_t$ (which correspond to right children in tree structures) also depend on the outcome of the fork and join variables, but are defined in terms of a side- and depth-specific sub-model in every case:%
\footnote{
Here $\phi_{\dvm} = \indicator{\bv^{1..\dvm}_\tv\!=\!\bv^{1..\dvm}_{\tv\dec1}}$,
     $\psi_{\dvm} = \indicator{\bv^{\dvm\inc1..\DV}_\tv\!=\!\bot}$,
and again, $\dvm \!=\! \max_{\dv}\{\bv^{\dv}_{\tv\dec1} \!\neq\! \bot\}$.
}
\begin{align}
\modelcondprob{\theta_\mathrm{B}} {\bv^{1..\DV}_\tv} {\qv_{\tv\dec1}\ \fv_\tv\ \pv_\tv\ \jv_\tv\ \av^{1..\DV}_\tv} \defeq 
\notag\\
  &\hspace{-4.8cm}\begin{cases}
    \phi_{\dvm\dec2} \cdot
    \modelcondprob{ \theta_{\mathrm{B},\mathrm{R},\dvm\dec1} }{ b^{\dvm\dec1}_t }{ b^{\dvm\dec1}_{t\dec1}\, a^{\dvm}_{t\dec1}}
    \!\!\!\!\!&\cdot\ \psi_{\dvm\inc0}
      \ \ \text{ if } f_t,\!j_t\!=\!0,1
    \\
    \phi_{\dvm\dec1} \cdot
    \modelcondprob{ \theta_{\mathrm{B},\mathrm{L},\dvm} }{ b^{\dvm}_t }{ a^{\dvm}_t\ a^{\dvm}_{t\dec1} } 
    \!\!\!\!&\cdot\ \psi_{\dvm\inc1}
      \ \ \text{ if } f_t,\!j_t\!=\!0,0
    \\
    \phi_{\dvm\dec1} \cdot
    \modelcondprob{ \theta_{\mathrm{B},\mathrm{R},\dvm} }{ b^{\dvm}_t }{ b^{\dvm}_{t\dec1}\ p_t}
    \!\!\!\!&\cdot\ \psi_{\dvm\inc1}
      \ \ \text{ if } f_t,\!j_t\!=\!1,1
    \\
    \phi_{\dvm\dec0} \cdot
    \modelcondprob{ \theta_{\mathrm{B},\mathrm{L},\dvm\inc1} }{ b^{\dvm\inc1}_t }{ a^{\dvm\inc1}_t\ p_t }  
    \!\!\!\!&\cdot\ \psi_{\dvm\inc2}
      \ \ \text{ if } f_t,\!j_t\!=\!1,0
  \end{cases}
\label{eqn:b}
\end{align}

In a sequence model inducer like \newcite{Shain2016}, these depth-specific models are assumed to be independent of each other and fit with a Gibbs sampler, 
backward sampling hidden variable sequences from forward distributions using this compiled transition model~$\MV$ \cite{Carter1996}, 
then counting individual sub-model outcomes from sampled hidden variable sequences, 
then resampling each sub-model using these counts with Dirichlet priors over $\av$, $\bv$, and $\pv$ models and Beta priors over $\fv$ and $\jv$ models, 
then re-compiling these resampled models into a new~$\MV$.

\def\KV{K}
\def\WV{W}
However, note that with $\KV$ category labels this model contains $\DV\KV^2+3\DV\KV^3$ separate parameters for preterminal categories and top and bottom categories of derivation fragments at every depth level, each of which can be independently learned by the Gibbs sampler.
Although this allows the hierarchical sequence model to learn grammars that are more expressive than PCFGs, the search space is several times larger than the $\KV^3$ space of PCFG nonterminal expansions.
The model described in this paper instead induces a PCFG and derives sequence model distributions from the PCFG, which has fewer parameters, and thus strictly reduces the search space of the model.

\section{The DB-PCFG Model}
\label{sect:model}

Unlike \newcite{Shain2016}, the depth-bounded probabilistic context-free grammar (DB-PCFG) model described in this paper directly induces a PCFG and then deterministically derives the parameters of a probabilistic left-corner parser from this single source. 
This derivation is based on an existing derivation of probabilistic left-corner parser models from PCFGs \cite{VanSchijndel2013}, which was developed in a supervised parsing model, here adapted to run more efficiently within a larger unsupervised grammar induction model.%
\footnote{
More specifically, the derivation differs from that of \newcite{VanSchijndel2013} in that it removes terminal symbols from conditional dependencies of models over fork and join decisions and top and bottom category labels, substantially reducing the size of the derived model that must be run during induction.
}

\def\GV{\mathbf{G}}
\def\cv{c}
A PCFG can be defined in Chomsky normal form as a matrix~$\GV$ of binary rule probabilities with one row for each of~$\KV$ parent symbols~$\cv$ and one column for each of $\KV^2{+}\WV$ combinations of left and right child symbols~$\av$ and~$\bv$, 
which can be pairs of nonterminals or observed words from vocabulary~$\WV$ followed by null symbols $\bot$:%
%
\footnote{
This definition assumes a Kronecker delta function $\delta_i$, defined as a vector with value one at index $i$ and zeros everywhere else, 
and a Kronecker product $\mathbf{M} \otimes \mathbf{N}$ over matrices~$\mathbf{M}$ and~$\mathbf{N}$, which tiles copies of $\mathbf{N}$ weighted by values in $\mathbf{M}$ as follows:
\begin{equation}
\mathbf{M} \otimes \mathbf{N} =
\left[ \begin{array}{ccc}
  \mathbf{M}_{[1,1]} \, \mathbf{N} & \mathbf{M}_{[1,2]} \, \mathbf{N} & \cdots \\
  \mathbf{M}_{[2,1]} \, \mathbf{N} & \mathbf{M}_{[2,2]} \, \mathbf{N} & \cdots \\
  \vdots                           & \vdots                           & \ddots
\end{array} \right]
\tag{1'}
\end{equation}
The Kronecker product specializes to vectors as single-column matrices, generating vectors that contain the products of all combinations of elements in the operand vectors.
}
\begin{equation}
\GV = \sum_{\av,\bv,\cv} \condprob{ \cv \rightarrow \av \ \bv }{ \cv } \ \delta_\cv \ (\delta_\av \otimes \delta_\bv)^\top
\end{equation}

\def\sv{s}
A depth-bounded grammar is a set of side- and depth-specific distributions:
\begin{equation}
\GV_\DV = \{ \GV_{\sv,\dv} \mid \sv\in\{\mathrm{L},\mathrm{R}\}, \dv\in\{1..\DV\} \}
\end{equation}
The posterior probability of a depth-bounded model~$\GV_\DV$ given a corpus (sequence) of words~$\wv_{1..\TV}$ is proportional to the product of a likelihood and a prior:
\begin{equation}
\condprob{\GV_\DV}{\wv_{1..\TV}} \propto \condprob{\wv_{1..\TV}}{\GV_\DV} \cdot \prob{\GV_\DV}
\end{equation}

The likelihood is defined as a marginal over bounded PCFG trees~$\tau$ of the probability of that tree given the grammar times the product of the probability of the word at each time step or token index~$\tv$ given this tree:%
\footnote{
This notation assumes the observed data~$\wv_{1..\TV}$ is a single long sequence of words, and the hidden variable~$\tau$ is a single large but depth-bounded tree structure (e.g.\ a right-branching discourse structure).
Since the implementation is incremental, segmentation decisions may indeed be treated as hidden variables in~$\tau$, but the experiments described in Section~\ref{sect:eval} are run on sentence-segmented input.
}
\begin{equation}
\condprob{\wv_{1..\TV}}{\GV_\DV}
=
\sum_\tau \condprob{\tau}{\GV_\DV} \cdot \prod_\tv \condprob{\wv_{\tv}}{\tau}
\end{equation}
The probability of each tree is defined to be the product of the probabilities of each of its branches:%
\footnote{
Here, $\eta$ is a node address, with left child $\eta0$ and right child $\eta1$, or with right child equal to $\bot$ if unary.
}
\begin{equation}
\condprob{\tau}{\GV_\DV} = \prod_{\tau_\eta\in\tau} \modelcondprob{\GV_\DV}{\tau_\eta \rightarrow \tau_{\eta0}\, \tau_{\eta1}}{\tau_\eta}
\end{equation}

The probability~$\prob{\GV_\DV}$ is itself an integral over the product of a deterministic transform~$\phi$ from an unbounded grammar to a bounded grammar~\mbox{$\condprob{\GV_\DV}{\GV} = \indicator{\GV_\DV = \phi(\GV)}$} and a prior over unbounded grammars~$\prob{\GV}$:
\begin{equation}
\prob{\GV_\DV} = \int \condprob{\GV_\DV}{\GV} \cdot \prob{\GV} \cdot d \GV
\end{equation}
Distributions~$\prob{\GV}$ for each nonterminal symbol (rows) within this unbounded grammar can then be sampled from a Dirichlet distribution with a symmetric parameter~$\beta$:
\begin{equation}
\GV \sim \mathrm{Dirichlet}( \beta )
\end{equation}
which then yields a corresponding transformed sample in $\prob{\GV_\DV}$ for corresponding nonterminals.
Note that this model is different than that of \newcite{Shain2016}, who induce a hierarchical HMM directly.

A depth-specific grammar~$\GV_\DV$ is (deterministically) derived from~$\GV$ via transform~$\phi$ with probabilities for expansions constrained to and renormalized over only those outcomes that yield terminals within a particular depth bound~$\DV$.
This depth-bounded grammar is then used to derive left-corner expectations (anticipated counts of categories appearing as left descendants of other categories), and ultimately the parameters of the depth-bounded left-corner parser defined in Section~\ref{sect:bkgd}.
Counts for~$\GV$ are then obtained from sampled hidden state sequences, and rows of~$\GV$ are then directly sampled from the posterior updated by these counts.

\subsection{Depth-bounded grammar}

In order to ensure the bounded version of~$\GV$ is a consistent probability model, it must be renormalized in transform~$\phi$ to assign a probability of zero to any derivation that exceeds its depth bound~$\DV$.
For example, if $\DV=2$, then it is not possible to expand a left sibling at depth 2 to anything other than a lexical item, so the probability of any non-lexical expansion must be removed from the depth-bounded model, and the probabilities of all remaining outcomes must be renormalized to a new total without this probability.
\def\hv{\mathbf{h}}
\def\sv{s}
\def\iv{i}
Following \newcite{VanSchijndel2013}, this can be done by iteratively defining a side- and depth-specific containment likelihood~$\hv^{(\iv)}_{\sv,\dv}$
for left- or right-side siblings~$\sv \in \{\mathrm{L},\mathrm{R}\}$
at depth~$\dv \in \{1..\DV\}$
\def\IV{I}
at each iteration~$\iv \in \{1..\IV\}$,%
\footnote{
Experiments described in this article use $\IV=20$ following observations of convergence at this point in supervised parsing.
}
as a vector with one row for each nonterminal or terminal symbol (or null symbol $\bot$) in~$\GV$, containing the probability of each symbol generating a complete yield within depth~$\dv$ as an $\sv$-side sibling:
\begin{subequations}
\begin{align}
\hv^{(0)}_{\sv,\dv}
 &=
  \mathbf{0}
\\
\hv^{(\iv)}_{\mathrm{L},\dv}
 &= 
  \begin{cases}
      \GV \, ( \mathbf{1} \otimes \delta_\bot 
                 \ + \ \hv^{(\iv-1)}_{\mathrm{L},\dv} \otimes \hv^{(\iv-1)}_{\mathrm{R},\dv} )
     & \!\!\text{if } \dv \leq \DV+1
   \\
   \mathbf{0}
     & \!\!\text{if } \dv > \DV+1
  \end{cases}
\\
\hv^{(\iv)}_{\mathrm{R},\dv}
 &= 
  \begin{cases}
   \delta_{\mathrm{T}}
     & \!\!\text{if } \dv = 0
   \\
   \GV \, ( \mathbf{1} \otimes \delta_\bot 
              \ + \ \hv^{(\iv-1)}_{\mathrm{L},\dv+1} \otimes \hv^{(\iv-1)}_{\mathrm{R},\dv} )
     & \!\!\text{if } 0 < \dv \leq \DV
   \\
   \mathbf{0}
     & \!\!\text{if } \dv > \DV
  \end{cases}
\end{align}
\end{subequations}
where `T' is a top-level category label at depth zero.

A depth-bounded grammar~$\GV_{\sv,\dv}$ can then be defined to be the original grammar~$\GV$ reweighted and renormalized by this containment likelihood:%
\footnote{
where~$\diag(\mathbf{v})$ is a diagonalization of a vector~$\mathbf{v}$:
\begin{equation}
\diag( \mathbf{v} ) = \left[ \begin{array}{ccc} \mathbf{v}_{[1]} & 0                & \cdots \\
                                                0                & \mathbf{v}_{[2]} &        \\
                                                \vdots           &                  & \ddots \end{array} \right]
\tag{2'}
\end{equation}
}
\begin{subequations}
\begin{align}
\GV_{\mathrm{L},\dv}
 &= \frac{ \GV \, \diag( \mathbf{1} \otimes \delta_\bot
                           \ + \ \hv^{(\IV)}_{\mathrm{L},\dv} \otimes \hv^{(\IV)}_{\mathrm{R},\dv} ) }
         { \hv^{(\IV)}_{\mathrm{L},\dv} }
\\
\GV_{\mathrm{R},\dv}
 &= \frac{ \GV \, \diag( \mathbf{1} \otimes \delta_\bot
                           \ + \ \hv^{(\IV)}_{\mathrm{L},\dv+1} \otimes \hv^{(\IV)}_{\mathrm{R},\dv} ) }
         { \hv^{(\IV)}_{\mathrm{R},\dv} }
\end{align}
\end{subequations}
This renormalization ensures the depth-bounded model is consistent.
Moreover, this distinction between a learned {\em unbounded} grammar~$\GV$ and a derived {\em bounded} grammar~$\GV_{s,d}$ which is used to derive a parsing model may be regarded as an instance of Chomsky's \shortcite{Chomsky1965} distinction between linguistic competence and performance.

\def\EV{\mathbf{E}}
The side- and depth-specific grammar can then be used to define expected counts of categories occurring as left descendants (or `left corners') of right-sibling ancestors:
\begin{subequations}
\begin{align}
\EV^{(1)}_\dv   &= \GV_{\mathrm{R},\dv} \, (\diag(\mathbf{1}) \otimes \mathbf{1})
\\
\EV^{(\iv)}_\dv &= \EV^{(\iv-1)}_\dv \, \GV_{\mathrm{L},\dv} \, (\diag(\mathbf{1}) \otimes \mathbf{1})
\\
\EV^+_\dv       &= \textstyle\sum^{\IV}_{\iv=1} \EV^{(\iv)}_\dv
\end{align}
\end{subequations}
This left-corner expectation will be used to estimate the marginalized probability over all grammar rule expansions between derivation fragments, which must traverse an unknown number of left children of some right-sibling ancestor.

\subsection{Depth-bounded parsing}
\label{sect:parsing}

Again following \newcite{VanSchijndel2013}, the fork and join decision, and the preterminal, top and bottom category label sub-models described in Section~\ref{sect:bkgd} can now be defined in terms of these side-and depth-specific grammars~$\GV_{\sv,\dv}$ and depth-specific left-corner expectations~$\EV^+_\dv$.

First, probabilities for no-fork and yes-fork outcomes below some bottom sign of category~$\bv$ at depth~$\dv$ are defined as the normalized probabilities, respectively, of any lexical expansion of a right sibling~$\bv$ at depth~$\dv$, and of any lexical expansion following any number of left child expansions from~$\bv$ at depth~$\dv$:
\begin{subequations}
\begin{align}
\modelcondprob{ \theta_{\mathrm{F},\dv} }{ 0 }{ \bv }
 &= \frac{ {\delta_\bv}^\top                \GV_{\mathrm{R},\dv} \, ( \mathbf{1} \otimes \delta_\bot ) }
         { {\delta_\bv}^\top ( \GV_{\mathrm{R},\dv} + \EV^+_\dv \, \GV_{\mathrm{L},\dv} ) \, ( \mathbf{1} \otimes \delta_\bot ) }
\\
\modelcondprob{ \theta_{\mathrm{F},\dv} }{ 1 }{ \bv }
 &= \frac{ {\delta_\bv}^\top   \EV^+_\dv \, \GV_{\mathrm{L},\dv} \, ( \mathbf{1} \otimes \delta_\bot ) }
         { {\delta_\bv}^\top ( \GV_{\mathrm{R},\dv} + \EV^+_\dv \, \GV_{\mathrm{L},\dv} ) \, ( \mathbf{1} \otimes \delta_\bot ) }\!
\end{align}
\end{subequations}

The probability of a preterminal~$\pv$ given a bottom category~$\bv$ is simply a normalized left-corner expected count of~$\pv$ under~$\bv$:
\begin{equation}
\modelcondprob{\theta_{\mathrm{P},\dv}} {\pv} {\bv}
 \defeq
  \frac{ {\delta_{\bv}}^\top \, \EV^+_\dv \, \delta_{\pv} }
       { {\delta_{\bv}}^\top \, \EV^+_\dv \, \mathbf{1}   }
\end{equation}

\def\cv{c}
Yes-join and no-join probabilities below bottom sign~$\bv$ and above top sign~$\av$ at depth~$\dv$ are then defined similarly to fork probabilities, as the normalized probabilities, respectively, of an expansion to left child~$\av$ of a right sibling~$\bv$ at depth~$\dv$, and of an expansion to left child~$\av$ following any number of left child expansions from~$\bv$ at depth~$\dv$:
\begin{subequations}
\begin{align}
\!\modelcondprob{ \theta_{\mathrm{J},\dv} }{ 1 }{ \bv \, \av }
 &= \frac{ {\delta_\bv}^\top                \GV_{\mathrm{R},\dv} \, ( \delta_\av \otimes \mathbf{1} ) }
         { {\delta_\bv}^\top ( \GV_{\mathrm{R},\dv} + \EV^+_\dv \, \GV_{\mathrm{L},\dv} ) \, ( \delta_\av \otimes \mathbf{1} ) }
\\
\!\modelcondprob{ \theta_{\mathrm{J},\dv} }{ 0 }{ \bv \, \av }
 &= \frac{ {\delta_\bv}^\top   \EV^+_\dv \, \GV_{\mathrm{L},\dv} \, ( \delta_\av \otimes \mathbf{1} ) }
         { {\delta_\bv}^\top ( \GV_{\mathrm{R},\dv} + \EV^+_\dv \, \GV_{\mathrm{L},\dv} ) \, ( \delta_\av \otimes \mathbf{1} ) }
\end{align}
\end{subequations}

The distribution over category labels for top signs~$\av$ above some top sign of category~$\cv$ and below a bottom sign of category~$\bv$ at depth~$\dv$ is defined as the normalized distribution over category labels following a chain of left children expanding from~$\bv$ which then expands to have a left child of category~$\cv$:
\begin{equation}
\modelcondprob{ \theta_{\mathrm{A},\dv} }{ \av }{ \bv\ \cv} =
 \frac{ {\delta_\bv}^\top \EV^+_\dv \, \diag(\delta_\av) \, \GV_{\mathrm{L},\dv} \, (\delta_\cv \otimes \mathbf{1}) }
      { {\delta_\bv}^\top \EV^+_\dv \, \diag(\mathbf{1}) \, \GV_{\mathrm{L},\dv} \, (\delta_\cv \otimes \mathbf{1}) }
\end{equation}

The distribution over category labels for bottom signs~$\bv$ below some sign~$\av$ and sibling of top sign~$\cv$ is then defined as the normalized distribution over right children of grammar rules expanding from~$\av$ to~$\cv$ followed by~$\bv$:
\begin{equation}
\modelcondprob{ \theta_{\mathrm{B},\sv,\dv} }{ \bv }{ \av\ \cv} =
 \frac{ {\delta_\av}^\top \GV_{\mathrm{\sv},\dv} \, (\delta_\cv \otimes \delta_\bv) }
      { {\delta_\av}^\top \GV_{\mathrm{\sv},\dv} \, (\delta_\cv \otimes \mathbf{1}) }
\end{equation}

\def\LV{\mathbf{L}}
Finally, a lexical observation model~$\LV$ is defined as a matrix of unary rule probabilities with one row for each combination of store state and preterminal symbol and one column for each observation symbol:
\begin{equation}
\LV = \mathbf{1} \otimes \GV \, ( \diag(\mathbf{1}) \otimes \delta_\bot )
\end{equation}

\subsection{Gibbs sampling}

Grammar induction in this model then follows a forward-filtering backward-sampling algorithm~\cite{Carter1996}.
\def\vv{\mathbf{v}}
This algorithm first computes a forward distribution~$\vv_\tv$ over hidden states at each time step~$\tv$ from an initial value $\bot$:
\begin{subequations}
\begin{align}
{ \vv_0 }^\top   &= { \delta_\bot }^\top
\\
{ \vv_\tv }^\top &= { \vv_{\tv-1} }^\top \, \MV \, \diag(\LV \, \delta_{\wv_\tv})
\end{align}
\end{subequations}
The algorithm then samples hidden states backward from a multinomial distribution given the previously sampled state~$\qv_{\tv+1}$ at time step~$\tv{+}1$ (assuming input parameters to the multinomial function are normalized):
\begin{equation}
\!\qv_\tv \sim \multinom( \, \diag( \vv_\tv ) \, \MV \, \diag(\LV \, \delta_{\wv_{\tv\inc1}}) \, \delta_{ \qv_{\tv+1} } \, )
\end{equation}

Grammar rule applications~$\CV$ are then counted from these sampled sequences:%
\footnote{
Again, $\dvm \!=\! \max_{\dv}\{\av^{\dv}_{\tv\dec1} \!\neq\! \bot\}$.
}
\begin{align}
\CV &= \sum_\tv
  \begin{cases}
  \delta_{\bv^{\dvm\dec1}_{\tv\dec1}} \, ( \delta_{\av^{\dvm}_{\tv\dec1}} \otimes \delta_{\bv^{\dvm\dec1}_\tv} )^\top
    & \!\!\!\text{if } \fv_\tv,\jv_\tv = 0,1
  \\
  \delta_{\av^{\dvm}_\tv} \, ( \delta_{\av^{\dvm}_{\tv\dec1}} \otimes \delta_{\bv^{\dvm}_\tv} )^\top
    & \!\!\!\text{if } \fv_\tv,\jv_\tv = 0,0
  \\
  \delta_{\bv^{\dvm}_{\tv\dec1}} \, ( \delta_{\pv_{\tv}} \otimes \delta_{\bv^{\dvm}_\tv} )^\top
    & \!\!\!\text{if } \fv_\tv,\jv_\tv = 1,1
  \\
  \delta_{\av^{\dvm\inc1}_\tv} \, ( \delta_{\pv_{\tv}} \otimes \delta_{\bv^{\dvm\inc1}_\tv} )^\top
    & \!\!\!\text{if } \fv_\tv,\jv_\tv = 1,0
  \end{cases}
\notag\\
 &+ \sum_\tv \delta_{\pv_\tv} \, ( \delta_{\wv_\tv} \otimes \delta_\bot )^\top
\end{align}
and a new grammar~$\GV$ is sampled from a Dirichlet distribution with counts~$\CV$ and a symmetric hyper-parameter~$\beta$ as parameters:
\begin{equation}
\GV \sim \dirich( \, \CV + \beta \, )
\end{equation}
This grammar is then used to define transition and lexical models~$\MV$ and~$\LV$ as defined in Sections~\ref{sect:bkgd} through~\ref{sect:parsing} to complete the cycle.

\subsection{Model hyper-parameters and priors}

There are three hyper-parameters in the model.
$K$ is the number of non-terminal categories in the grammar $\GV$, $\DV$ is the maximum depth, and $\beta$ is the parameter for the symmetric Dirichlet prior over multinomial distributions in the grammar $\GV$.

As seen from the previous subsection, the prior is over all possible rules in an unbounded PCFG grammar.
Because the number of non-terminal categories of the unbounded PCFG grammar is given as a hyper-parameter, the number of rules in the grammar is always known.
It is possible to use non-parametric priors over the number of non-terminal categories, however due to the need to dynamically mitigate the computational complexity of filtering and sampling using arbitrarily large category sets, this is left for future work.

\section{Evaluation}
\label{sect:eval}

The DB-PCFG model described in Section~\ref{sect:model} is evaluated first on synthetic data to determine whether it can reliably learn a recursive grammar from data with a known optimum solution, and to determine the hyper-parameter value for $\beta$ for doing so.
Two experiments on natural data are then carried out.
First, the model is run on natural data from the Adam and Eve parts of the CHILDES corpus \cite{macwhinney00} to compare with other grammar induction systems on a human-like acquisition task.
Then data from the Wall Street Journal section of the Penn Treebank \cite{marcusetal93} is used for further comparison 
in a domain for which competing systems are optimized. The competing systems include UPPARSE \cite{Ponvert2011}%
\footnote{\url{https://github.com/eponvert/upparse}},
CCL \cite{Seginer2007a}%
\footnote{\url{https://github.com/DrDub/cclparser}},
BMMM+DMV with undirected dependency features \cite{Christodoulopoulos}%
\footnote{BMMM:\url{https://github.com/christos-c/bmmm} \\DMV:\url{https://code.google.com/archive/p/pr-toolkit/}}
and UHHMM \cite{Shain2016}.%
\footnote{\url{https://github.com/tmills/uhhmm/tree/coling16}}

For the natural language datasets, the variously parametrized DB-PCFG systems%
\footnote{The most complex configuration that would run on available GPUs was $D{=}2,K{=}15$.
Analysis of full WSJ \cite{Schuler2010} shows 47.38\% of sentences require depth 2, 38.32\% require depth 3 and 6.26\% require depth 4.}
are first validated on a development set, and the optimal system is then run until convergence with the chosen hyperparameters on the test set. In development experiments, the log-likelihood of the dataset plateaus usually after 500 iterations. The system is therefore run at least 500 iterations in all test set experiments, with one iteration being a full cycle of Gibbs sampling.  The system is then checked to see whether the log-likelihood has plateaued, and halted if it has. 

The DB-PCFG model assigns trees sampled from conditional posteriors to all sentences in a dataset in every iteration as part of the inference. The system is further allowed to run at least 250 iterations after convergence and proposed parses are chosen from the iteration with the greatest log-likelihood after convergence. However, once the system reaches convergence, the evaluation scores of parses from different iterations post-convergence appear to differ very little.

\subsection{Synthetic data}

Following \newcite{Liang2009} and \newcite{Scicluna}, an initial set of experiments on synthetic data are used to investigate basic properties of the model---in particular:
\begin{enumerate}
\item whether the model is balanced or biased in favor of left- or right-branching solutions,
\item whether the model is able to posit recursive structure in appropriate places, and
\item what hyper-parameters enable the model to find optimal modes more quickly.
\end{enumerate}

\begin{figure}
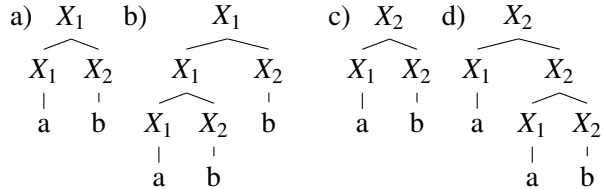

\tikzset{level distance=1.5\baselineskip}
a)
\!\!\!\!\Tree [.$X_1$ [.$X_1$ a ] [.$X_2$ b ] ]
b)
\!\!\!\!\Tree [.$X_1$ [.$X_1$ [.$X_1$ a ] [.$X_2$ b ] ] [.$X_2$ b ] ]
\ \ \ \ \ \ 
c)
\!\!\!\!\Tree [.$X_2$ [.$X_1$ a ] [.$X_2$ b ] ]
d)
\!\!\!\!\Tree [.$X_2$ [.$X_1$ a ] [.$X_2$ [.$X_1$ a ] [.$X_2$ b ] ] ]

\caption{
Synthetic left-branching (a,b) and right-branching (c,d) datasets.
}
\label{fig:LRsynthtrees}
\end{figure}

The risk of bias in branching structure is important because it might unfairly inflate induction results on languages like English, which are heavily right branching.
In order to assess its bias, the model is evaluated on two synthetic datasets, each consisting of 200 sentences.
The first dataset is a left-branching corpus, which consists of 100 sentences of the form $a\, b$ and 100 sentences of the form $a\, b\, b$ 
, with optimal tree structures as shown in Figure~\ref{fig:LRsynthtrees} (a) and (b).
The second dataset is a right-branching corpus, which consists of 100 sentences of the form $a\, b$ and 100 sentences of the form $a\, a\, b$
, with optimal tree structures as shown in Figure~\ref{fig:LRsynthtrees} (c) and (d).
Results show both structures (and both corresponding grammars) are learnable by the model, and result in approximately the same log likelihood.
These synthetic datasets are also used to tune the~$\beta$ hyper-parameter of the model (as defined in Section~\ref{sect:model}) to enable it to find optimal modes more quickly.
The resulting $\beta$ setting of 0.2 is then used in induction on the CHILDES and Penn Treebank corpora.

\begin{figure}
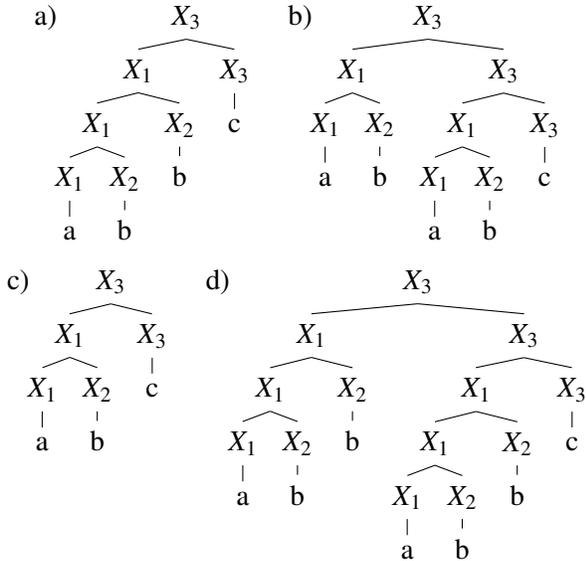

\tikzset{level distance=1.5\baselineskip}
\begin{center}
a)\!\!\!\!
\Tree [.$X_3$ [.$X_1$ [.$X_1$ [.$X_1$ a ] [.$X_2$ b ] ] [.$X_2$ b ] ] [.$X_3$ c ] ]
\ \ \ \ 
b)\!\!\!\!
\Tree [.$X_3$ [.$X_1$ [.$X_1$ a ] [.$X_2$ b ] ] [.$X_3$ [.$X_1$ [.$X_1$ a ] [.$X_2$ b ] ] [.$X_3$ c ] ] ]

\smallskip
c)\!\!\!\!
\Tree [.$X_3$ [.$X_1$ [.$X_1$ a ] [.$X_2$ b ] ] [.$X_3$ c ] ]
\ \ \ \ 
d)\!\!\!\!
\Tree [.$X_3$ [.$X_1$ [.$X_1$ [.$X_1$ a ] [.$X_2$ b ] ] [.$X_2$ b ] ] [.$X_3$ [.$X_1$ [.$X_1$ [.$X_1$ a ] [.$X_2$ b ] ] [.$X_2$ b ] ] [.$X_3$ c ] ] ]
\end{center}
\caption{
Synthetic center-embedding structure.
Note that tree structures (b) and (d) have depth 2 because they have complex sub-trees spanning $a\, b$ and $a\, b\, b$, respectively, embedded in the center of the yield of their roots.
}
\label{fig:CEsynthtrees}
\end{figure}

After validating that the model is not biased, the model is also evaluated on a synthetic center-embedding corpus consisting of 50 sentences each of the form $a\, b\, c$; $a\, b\, b\, c$; $a\, b\, a\, b\, c$; and $a\, b\, b\, a\, b\, b\, c$, which has optimal tree structures as shown in Figure~\ref{fig:CEsynthtrees}.%
\footnote{
Here, in order to more closely resemble natural language input, tokens $a$, $b$, and $c$ are randomly chosen uniformly from $\{a_1, \dots, a_{50}\}$, $\{b_1, \dots, b_{50}\}$ and $\{c_1, \dots, c_{50}\}$, respectively.
}
Note that the (b) and (d) trees have depth 2 because they each have a complex sub-tree spanning $a\, b$ and $a\, b\, b$ embedded in the center of the yield of the root.
Results show the model is capable of learning depth 2 (recursive) grammars.

Finally, as a gauge of the complexity of this task, results of the model described in this paper are compared with those of other grammar induction models on the center-embedding dataset.
In this experiment, all models are assigned hyper-parameters matching the optimal solution.
The DB-PCFG is run with $K{=}5$ and $D{=}2$ and $\beta{=}0.2$ for all priors,
the BMMM+DMV \cite{Christodoulopoulos} is run with 3 preterminal categories,
and the UHHMM model is run with 2 active states, 4 awaited states and 3 parts of speech.%
\footnote{It is not possible to use just 2 awaited states, which is the gold setting, since the UHHMM system errors out when the number of categories is small.}
Table \ref{tab:synthetic2} shows the PARSEVAL scores for parsed trees using the learned grammar from each unsupervised system.
Only the DB-PCFG model is able to recognize the correct tree structures and the correct category labels on this dataset, showing the task is indeed a robust challenge.
This suggests that hyper-parameters optimized on this dataset may be portable to natural data.

\begin{table}
\centering
\begin{tabular}{c|c|c|c}
\hline
System	&	Precision	&	Recall	&	F1	\\
\hline
CCL		&	83.2	&	71.1	&	76.7	\\
UPPARSE	&	91.4		&	80.7	&	85.7	\\
UHHMM	&	37.7	&	37.7	&	37.7	\\
BMMM+DMV	&	99.2	&	83.2	&	90.5	\\
\hline
DB-PCFG	&	\textbf{100.0}		&	\textbf{100.0}	&	\textbf{100.0}	\\
\hline
\end{tabular}
\caption{The performance scores of unlabeled parse evaluation of different systems on synthetic data. }
\label{tab:synthetic2}
\end{table}

\begin{table}
\centering
\begin{tabular}{c|c|c|c}
\hline
Hyperparameters	&	Precision	&	Recall	&	F1	\\
\hline
D1K15		&	\textbf{57.1}	&	\textbf{70.7}	&	\textbf{63.2}	\\
D1K30	&	52.8	&	65.4	&	58.5	\\
D1K45	&	44.4	&	54.9	&	49.1	\\
D2K15	&	44.0	&	54.5	&	48.7	\\

\hline
\end{tabular}
\caption{PARSEVAL results of different hyperparameter settings for the DB-PCFG system on the Adam dataset. Hyperparameter $D$ is the number of possible depths, and $K$ is the number of non-terminals. }
\label{tab:adam}
\end{table}

\subsection{Child-directed speech corpus}

After setting the $\beta$ hyperparameter on synthetic datasets, the DB-PCFG model is evaluated on 14,251 sentences of transcribed child-directed speech from the Eve section of the Brown corpus of CHILDES \cite{macwhinney00}. 
Hyperparameters $D$ and $K$ are set to optimize performance on the Adam section of the Brown Corpus of CHILDES, which is about twice as long as Eve. 
Following \newcite{Seginer2007a}, \newcite{Ponvert2011} and \newcite{Shain2016}, these experiments leave all punctuation in the input for learning, then remove it in all evaluations on development and test data.

Model performance is evaluated against Penn Treebank style annotations of both Adam and Eve corpora \cite{Pearl2013}.
Table \ref{tab:adam} shows the PARSEVAL scores of the DB-PCFG system with different hyperparameters on the Adam corpus for development.%
The simplest configuration, D1K15 (depth 1 only with 15 non-terminal categories), obtains the best score, so this setting is applied to the test corpus, Eve.
Results of the $D{=}1,K{=}15$ DB-PCFG model on Eve are then compared against those of other grammar induction systems which use only raw text as input on the same corpus.
Following \newcite{Shain2016} the BMMM+DMV system is run for 10 iterations with 45 categories and its output is converted from dependency graphs to constituent trees \cite{Collins}.
The UHHMM system is run on the Eve corpus using settings in \newcite{Shain2016}, which also includes a post-process option to flatten trees (reported here as UHHMM-F).
%

\begin{table}
\centering
\begin{tabular}{c|c|c|c}
\hline
System				&	Precision	&	Recall	&	F1	\\
\hline
CCL					&	50.5		&	53.5	&	51.9	\\
UPPARSE				&	60.5		&	51.9	&	55.9	\\
UHHMM				&	55.5		&	69.3	&	61.7	\\
BMMM+DMV			&	63.5 &	63.3	&	63.4	\\
UHHMM-F	&	62.9		&	68.4	&	65.6	\\
\hline
DB-PCFG	&	\textbf{64.5}	&	\textbf{80.5}	&	$\textbf{71.6}^{**}$	\\
\hline
Right-branching	&	\textbf{68.7}	&	\textbf{85.8}	&	\textbf{76.3}	\\
\hline
\end{tabular}
\caption{PARSEVAL scores on Eve dataset for all competing systems. These are unlabeled precision, recall and F1 scores on constituent trees without punctuation. Both the right-branching baseline and the best performing system are in bold. (**: $p$ \textless~ 0.0001, permutation test) }
\label{tab:parseval}
\end{table}


Table \ref{tab:parseval} shows the PARSEVAL scores for all the competing systems on the Eve dataset.
The right-branching baseline is still the most accurate in terms of PARSEVAL scores, presumably because of the highly right-branching structure of child-directed speech in English.
%
The DB-PCFG system with only one memory depth and 15 non-terminal categories achieves the best performance in terms of F1 score and recall among all the competing systems, significantly outperforming other systems ($p$ \textless~0.0001, permutation test).%
%
%
\footnote{Resulting scores are better when applying \newcite{Shain2016} flattening to output binary-branching trees. For the $D{=}1$, $K{=}15$ model, precision and F1 can be raised to 70.31\% and 74.33\%. However, since the flattening is a heuristic which may not apply in all cases, these scores are not considered to be comparable results.}

The Eve corpus has about 5,000 sentences with more than one depth level, therefore one might expect a depth-two model to perform better than a depth-one model, but this is not true if only PARSEVAL scores are considered.
This issue will be revisited in the following section with the noun phrase discovery task.

\subsection{NP discovery on child-directed speech}

When humans acquire grammar, they do not only learn tree structures, they also learn category types: noun phrases, verb phrases, prepositional phrases, and where each type can and cannot occur.

Some of these category types --- in particular, noun phrases --- are fairly universal across languages, and may be useful in downstream tasks such as (unsupervised) named entity recognition.
The DB-PCFG and other models that can be made to produce category types are therefore evaluated on a noun phrase discovery task.

Two metrics are used for this evaluation.
First, the evaluation counts all constituents proposed by the candidate systems, and calculates recall against the gold annotation of noun phrases.
This metric is not affected by which branching paradigm the system is using and reveals more about the systems' performances.
This metric differs from that used by \newcite{Ponvert2011} in that this metric takes NPs at all levels in gold annotation into account, not just base NPs.%
\footnote{\newcite{Ponvert2011} define base NPs as NPs with no NP descendants, a restriction motivated by their particular task (chunking).}

\begin{table}
\centering
\begin{tabular}{c|c|c}
\hline
System	&	NP Recall	&	NP agg F1\\
\hline
CCL		&	35.5	&	-	\\
UPPARSE	&	69.1		&	-	\\
UHHMM	&	61.4	&	27.4	\\
BMMM+DMV	&	71.3		&	61.2	\\
\hline
DB-PCFG (D1K15)	&	75.7	&	28.7	\\
DB-PCFG (D1K30)	&	78.6	&	60.7\\
DB-PCFG (D1K45)	&	76.9	&	64.0	\\
DB-PCFG (D2K15)	&	\textbf{85.1}	&	\textbf{65.9}	\\
\hline
Right-branching	&	64.2	&	-	\\
\hline
\end{tabular}
\caption{Performances of different systems for noun phrase recall and aggregated F1 scores on the Eve dataset.}
\label{tab:np_recall}
\end{table}

\begin{table*}[t]
\centering
\begin{tabular}{c|c|c|c|c|c|c}
\hline
\multirow{2}{*}{System}	& \multicolumn{3}{c|}{WSJ10test} & \multicolumn{3}{c}{WSJ20test}		\\
\cline{2-7}
 & Precision	&	Recall	&	F1 & Precision	&	Recall	&	F1 \\
\hline
CCL					&	63.4	&	71.9	&	67.4	&	\textbf{60.1}	&	61.7	& $\textbf{60.9}^{**}$\\
UPPARSE				&	54.7	&	48.3	&	51.3	&	47.8	&	40.5 & 43.9 \\
UHHMM				&	49.1	&	63.4	&	55.3	&	-	&	-	& - \\
BMMM+DMV(K10)		&	36.2 &		40.6	&	38.2	&	25.3	&	29.0	& 27.0\\
UHHMM-F	&	57.1	&	54.4	&	55.7	&	-	&	-	& - \\
\hline
DB-PCFG (D2K15)	&	\textbf{64.5}	&	\textbf{82.6}	&	$\textbf{72.4}^{**}$	&	53.0	&	\textbf{70.5}	&	60.5 \\
\hline
Right-branching	&	55.1	&	70.5	&	61.8	&	41.5	&	55.3	& 47.4 \\
\hline
\end{tabular}
\caption{PARSEVAL scores for all competing systems on WSJ10 and WSJ20 test sets. These are unlabeled precision, recall and F1 scores on constituent trees without punctuation (**: $p$ \textless 0.0001, permutation test).}
\label{tab:wsj20}
\end{table*}

The second metric, for systems that produce category labels, calculates F1 scores of induced categories that can be mapped to noun phrases.
The first 4,000 sentences are used as the development set for learning mappings from induced category labels to phrase types.
The evaluation calculates precision, recall and F1 of all spans of proposed categories against the gold annotations of noun phrases in the development set, and aggregates the categories ranked by their precision scores so that the F1 score of the aggregated category is the highest on the development set.
The evaluation then calculates the F1 score of this aggregated category on the remainder of the dataset, excluding this development set.

The UHHMM system is the only competing system that is natively able to produce labels for proposed constituents.
BMMM+DMV does not produce constituents with labels by default, but can be evaluated using this metric by converting dependency graphs into constituent trees, then labeling each constituent with the part-of-speech tag of the head. For CCL and UPPARSE, the NP agg F1 scores are not reported because they do not produce labeled constituents.

Table \ref{tab:np_recall} shows the scores for all systems on the Eve dataset and four runs of the DB-PCFG system on these two evaluation metrics.
Surprisingly the $D{=}2$, $K{=}15$ model which has the lowest PARSEVAL scores is most accurate at discovering noun phrases.
It has the highest scores on both evaluation metrics.
The best model in terms of PARSEVAL scores, the $D{=}1$, $K{=}15$ DB-PCFG model, performs poorly among the DB-PCFG models, despite the fact that its NP Recall is higher than the competing systems. The low score of NP agg F1 of DB-PCFG at D1K15 shows a diffusion of induced syntactic categories when the model is trying to find a balance among labeling and branching decisions.
The UPPARSE system, which is proposed as a base NP chunker, is relatively poor at NP recall by this definition.

The right-branching baseline does not perform well in terms of NP recall.
This is mainly because noun phrases are often left children of some other constituent and the right branching model is unable to incorporate them into the syntactic structures of whole sentences.
Therefore although the right-branching model is the best model in terms of PARSEVAL scores, it is not helpful in terms of finding noun phrases.
%

\begin{table*}[t]
\centering
\begin{tabular}{c|c|c|c|c|c|c}
\hline
\multirow{2}{*}{System}	& \multicolumn{3}{c|}{WSJ10} & \multicolumn{3}{c}{WSJ40}		\\
\cline{2-7}
 & Precision	&	Recall	&	F1 & Precision	&	Recall	&	F1 \\
\hline
CCL					&	75.3	&	76.1	&	75.7	&	58.7	&	55.9	& 57.2 \\
UPPARSE				&	74.6	&	66.7	&	70.5	&	60.0	&	49.4 & 54.2 \\
\hline
DB-PCFG (D2K15)	&	65.5	&	83.6	&	73.4	&	47.0	&	63.6	&	54.1 \\
\hline
Right-branching	&	55.2	&	70.0	&	61.7	&	35.4	&	47.4	& 40.5 \\
\hline
\end{tabular}
\caption{Published PARSEVAL results for competing systems. Please see text for details as the systems are trained and evaluated differently.}
\label{tab:wsj40}
\end{table*}

\subsection{Penn Treebank}

To further facilitate direct comparison to previous work, we run experiments on sentences from the Penn Treebank~\cite{marcusetal93}. The first experiment uses the sentences from Wall Street Journal part of the Penn Treebank with at most 20 words  (WSJ20). The first half of the WSJ20 dataset is used as a development set (WSJ20dev) and the second half is used as a test set (WSJ20test). We also extract sentences in WSJ20test with at most 10 words from the proposed parses from all systems and report results on them (WSJ10test). WSJ20dev is used for finding the optimal hyperparameters for both DB-PCFG and BMMM-DMV systems.%
\footnote{Although UHHMM also needs tuning, in practice we find that this system is too inefficient to be tuned on a development set, and it requires too many resources when the hyperparameters become larger than used in previous work. We believe that further increasing the hyperparameters of UHHMM may lead to performance increase, but the released version is not scalable to larger values of these settings. We also do not report UHHMM on WSJ20test for the same scalabilty reason. The results of WSJ10test of UHHMM is induced with all WSJ10 sentences.}

Table \ref{tab:wsj20} shows the PARSEVAL scores of all systems. The right-branching baseline is relatively weak on these two datasets, mainly because formal writing is more complex and uses more non-right-branching structures (e.g., subjects with modifiers or parentheticals) than child-directed speech. For WSJ10test, both the DB-PCFG system and CCL are able to outperform the right branching baseline. The F1 difference between the best-performing previous-work system, CCL, and DB-PCFG is highly significant. For WSJ20test, again both CCL and DB-PCFG are above the right-branching baseline. The difference between the F scores of CCL and DB-PCFG is very small compared to WSJ10, however it is also significant.

It is possible that the DB-PCFG is being penalized for inducing fully binarized parse trees.
The accuracy of the DB-PCFG model is dominated by recall rather than precision, whereas CCL and other systems are more balanced.
This is an important distinction if it is assumed that phrase structure is binary \cite{Kayne1981,Larson1988}, in which case precision merely scores non-linguistic decisions about whether to suppress annotation of non-maximal projections.
However, since other systems are not optimized for recall, it would not be fair to use only recall as a comparison metric in this study.

Finally, Table \ref{tab:wsj40} shows the published results of different systems on WSJ. The CCL results come from \newcite{Seginer2007}, where the CCL system is trained with all sentences from WSJ, and evaluated on sentences with 40 words or fewer from WSJ (WSJ40) and WSJ10. The UPPARSE results come from \newcite{Ponvert2011}, where the UPPARSE system is trained using 00-21 sections of WSJ, and evaluated on section 23 and the WSJ10 subset of section 23. The DB-PCFG system uses hyperparameters optimized on the WSJ20dev set, and is evaluated on WSJ40 and WSJ10, both excluding WSJ20dev. The results are not directly comparable, but the results from the DB-PCFG system is competitive with the other systems, and numerically have the best recall scores.%

\section{Conclusion}

This paper describes a Bayesian Dirichlet model of depth-bounded PCFG induction.
Unlike earlier work this model implements depth bounds directly on PCFGs by derivation, reducing the search space of possible trees for input words without exploding the search space of parameters with multiple side- and depth-specific copies of each rule.
Results for this model on grammar acquisition from transcribed child-directed speech and newswire text exceed or are competitive with those of other models when evaluated on parse accuracy.
Moreover, grammars acquired from this model demonstrate a consistent use of category labels, something which has not been demonstrated by other acquisition models.

In addition to its practical merits, this model may offer some theoretical insight for linguists and other cognitive scientists.
First, the model does not assume any universals except independently motivated limits on working memory,
which may help address the question of whether universals are indeed necessary for grammar induction.
Second, the distinction this model draws between its learned {\em unbounded} grammar~$\GV$ and its derived {\em bounded} grammar~$\GV_\DV$ seems to align with Chomsky's \shortcite{Chomsky1965} distinction between competence and performance, and has the potential to offer some formal guidance to linguistic inquiry about both kinds of models.


\section*{Acknowledgments}
The authors would like to thank Cory Shain and William Bryce and the anonymous reviewers for their valuable input.
Computations for this project were partly run 
on the Ohio Supercomputer Center
\shortcite{OhioSupercomputerCenter1987}.
This research was funded by the Defense Advanced Research Projects Agency award HR0011-15-2-0022.
The content of the information does not necessarily reflect the position or the policy of the Government, and no official endorsement should be inferred.


\bibliography{Mendeley_tacl2017,tacl}
\bibliographystyle{acl2012}

\end{document}